\documentclass{article}

\usepackage{times}
\usepackage{graphicx} 
\usepackage{subfigure} 

\usepackage{natbib}
\usepackage{amsmath}
\usepackage{amssymb}
\usepackage{enumitem}

\usepackage{algorithm}
\usepackage{algorithmic}

\usepackage{hyperref}

\newcommand{\setS}{\mathcal{S}}

\usepackage[accepted]{icml2017} 

\icmltitlerunning{Sequence Modeling via Segmentations}

\begin{document} 

\twocolumn[
\icmltitle{Sequence Modeling via Segmentations}

\icmlsetsymbol{equal}{*}

\begin{icmlauthorlist}
\icmlauthor{Chong Wang}{msr}
\icmlauthor{Yining Wang}{cmu}
\icmlauthor{Po-Sen Huang}{msr}
\icmlauthor{Abdelrahman Mohamed}{amzn}
\icmlauthor{Dengyong Zhou}{msr}
\icmlauthor{Li Deng}{citadel}
\end{icmlauthorlist}

\icmlaffiliation{msr}{Microsoft Research}
\icmlaffiliation{cmu}{Carnegie Mellon University}
\icmlaffiliation{citadel}{Citadel Securities LLC}
\icmlaffiliation{amzn}{Amazon}

\icmlcorrespondingauthor{Chong Wang}{chowang@microsoft.com}

\icmlkeywords{boring formatting information, machine learning, ICML}

\vskip 0.3in
]

\printAffiliationsAndNotice{}  

\begin{abstract} 

Segmental structure is a common pattern in many types of sequences such as phrases in
human languages. In this paper, we present a probabilistic model for sequences 
via their segmentations.\footnote{The source code is available at \url{https://github.com/posenhuang/NPMT}.} 
The probability of a segmented sequence 
is calculated as the product of the probabilities of all its segments, 
where each segment is modeled using existing tools such as recurrent neural networks. 
Since the segmentation of a sequence is usually unknown in advance, 
we sum over all valid segmentations to obtain the final probability for the sequence. 
An efficient dynamic programming algorithm is developed for 
forward and backward computations without resorting to any approximation. 
We demonstrate our approach on text segmentation and speech recognition tasks. 
In addition to quantitative results, we also show that our approach can 
discover meaningful segments in their respective application contexts.

\end{abstract} 
\section{Introduction}
\label{sec:intro}


Segmental structure is a common pattern in many types of sequences, typically,  
phrases in human languages and letter combinations in phonotactics rules. For instances, 
\vspace{-3mm}
\begin{itemize}[noitemsep]
\item Phrase structure.  ``Machine learning is part of artificial 
intelligence''  $\rightarrow$ [Machine learning] [is] [part of] [artificial intelligence].
\item Phonotactics rules.  ``thought'' $\rightarrow$ [th][ou][ght].
\end{itemize}
\vspace{-3mm}
The words or letters in brackets ``[ ]'' are usually considered as 
meaningful segments for the original sequences. 
In this paper, we hope to incorporate this type 
of segmental structure information into sequence modeling.

Mathematically,  we are interested in constructing a conditional
probability distribution $p(y|x)$, where output $y$ is a sequence 
and input $x$ may or may not be a sequence. 
Suppose we have a segmented sequence. 
Then the probability of this  sequence
is calculated as the product of the probabilities of its segments, 
each of which is modeled using existing tools such as recurrent neural networks (RNNs), long-short term memory (LSTM)~\cite{Hochreiter:1997}, or gated recurrent 
units (GRU)~\cite{Chung:2014}.
When the segmentation for a sequence 
is unknown, we sum over the probabilities from all
valid segmentations. In the case that the input is also a sequence,  
we further need to sum over all feasible alignments
between inputs and output segmentations. This sounds complicated. Fortunately, we show
that both forward and backward computations can be tackled with a 
dynamic programming algorithm without resorting to any approximations. 


This paper is organized as follows. In Section~\ref{sec:model}, we
describe our mathematical model which constructs the probability 
distribution of a sequence via its segments, and discuss related work. 
In Section~\ref{sec:alg}, we present an
efficient dynamic programming algorithm for forward and backward
computations, and a beam search algorithm for decoding the output. 
Section~\ref{sec:exp} includes two case studies to
demonstrate the usefulness of our approach through both 
quantitative and qualitative results. We conclude this paper and
discuss future work in Section 5.


\section{Sequence modeling via segmentations}
\label{sec:model}
In this section, we present our formulation of sequence modeling via
segmentations. In our model, the output is always a sequence, while the
input may or may not be a sequence. We first consider the non-sequence
input case, and then move to the sequence input case. We then show how
to carry over information across segments when needed. Related work is
also discussed here.

\subsection{Case I: Mapping from non-sequence to sequence}
\label{subsec:non-seq}
\begin{figure}[t]
\begin{center}
\centerline{\includegraphics[width=0.83\columnwidth]{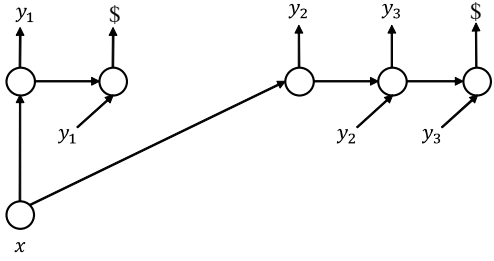}}
\vskip -0.22in
\caption{For Section~\ref{subsec:non-seq}. Given output $y_{1:3}$ and
its segmentation $a_1 = \{y_1, \$\}$ and $a_2 = \{y_2, y_3, \$ \}$,
input $x$ controls the initial states of both segments. Note that
$\pi(a_{1:t-1})$ is omitted here.}
\label{fig:non-seq-input}
\end{center}
\vskip -0.25in
\end{figure} 
Assume the input $x$ is a fixed-length vector.  Let the output
sequence be $y_{1:T}$. We are interested in modeling the probability 
$p(y_{1:T} |x)$ via the segmentations of $y_{1:T}$. Denote by
$\setS_y$ the set containing all valid segmentations of $y_{1:T}$.
Then for any segmentation $a_{1:\tau_a} \in \setS_{y}$, we have
$\pi(a_{1:\tau_a}) = y_{1:T}$, where $\pi(\cdot)$ is the concatenation
operator and $\tau_a$ is the number of segments in this segmentation.
For example, let $T=5$ and $\tau_a=3. $ Then one possible
$a_{1:\tau_a}$ could be like $a_{1:\tau_a} = \{\{y_1, \$\},\{y_2, y_3,
\$\}, \{y_4,y_5, \$\} \}$, where $\$$ denotes the end of a segment. Note that 
symbol $\$$ will be ignored in the concatenation operator $\pi(\cdot)$.
Empty segments, those containing only $\$$, are {\it not} permitted in
our setting. Note that while the number of distinct segments for a
length-$T$ sequence is $O(T^2)$, the number of distinct segmentations,
that is, $|\setS_y|$, is exponentially large.

Since the segmentation is unknown in advance, the probability of the
sequence $y_{1:T}$ is defined as the sum of the probabilities from all
the segmentations in $\setS_y$,
\begin{align}
  p(y_{1:T} |x) &\triangleq \sum_{a_{1:\tau_a} \in \setS_y} p(a_{1:\tau_a}|x) \nonumber \\ 
  &= \sum_{a_{1:\tau_a} \in \setS_y}
  \prod_{t=1}^{\tau_a} p(a_t|x, \pi(a_{1:t-1})), 
  \label{eq:model-1}
\end{align}
where $p(a_{1:\tau_a}|x)$ is the probability for segmentation
$a_{1:\tau_a}$ given input $x$, and $p(a_t|x, \pi(a_{1:t-1}))$ is the
probability for segment $a_t$ given input $x$ and the
concatenation of all previous segments $\pi(a_{1:t-1})$.
Figure~\ref{fig:non-seq-input} illustrates a possible relationship
between $x$ and $y_{1:T}$ given one particular segmentation.  We
choose to model the segment probability $p(a_t|x, \pi(a_{1:t-1}))$
using recurrent neural networks (RNNs), such as LSTM or GRU, with 
a softmax probability function. Input $x$ and concatenation
$\pi(a_{1:t-1})$ determine the initial state for this RNN. (All
segments' RNNs share the same network parameters.) However, since
$|\setS_y|$ is exponentially large, Eq.~\ref{eq:model-1} cannot be
directly computed. We defer the computational details to Section~\ref{sec:alg}. 

\subsection{Case II: Mapping from sequence to sequence}
\label{subsec:seq}
\begin{figure}[t]
\begin{center}
\centerline{\includegraphics[width=0.85\columnwidth]{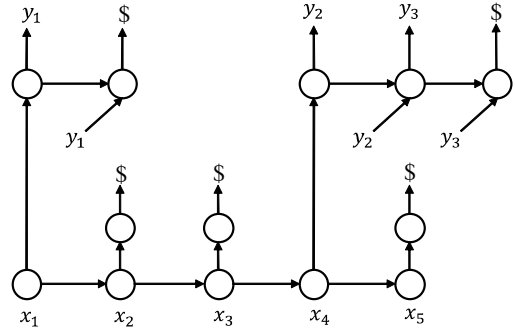}}
\vskip -0.22in
\caption{For Section~\ref{subsec:seq}. SWAN emits one particular
segmentation of $y_{1:T}$  with $x_1$ waking (emits $y_1$) and $x_4$
waking (emits $y_2$ and $y_3$) while $x_2$, $x_3$ and $x_5$ sleeping. SWAN
needs to consider all valid segmentations like this for $y_{1:T}$.}
\label{fig:wasm}
\end{center}
\vskip -0.25in
\end{figure}

Now we assume the input is also a sequence $x_{1:T'}$ and the output
remains as $y_{1:T}$. We make a {\it monotonic} alignment assumption---each
input element $x_t$ emits one segment $a_t$, which is then
concatenated as $\pi(a_{1:T'})$ to obtain $y_{1:T}$.  Different from
the case when the input is not a sequence, we allow empty
segments in the emission, i.e., $a_t = \{\$\}$ for some $t$, such that
any segmentation of $y_{1:T}$ will always consist of exactly $T'$
segments with possibly some empty ones. In other words, all valid
segmentations for the output is in set $\setS_y \triangleq
\{a_{1:T'}: \pi(a_{1:T'}) = y_{1:T}\}$. Since an input element can
choose to emit an empty segment, we name this particular method as
``{S}leep-{WA}ke {N}etworks'' (SWAN). See Figure~\ref{fig:wasm} for an
example of the emitted segmentation of $y_{1:T}$. 

Again, as in Eq.~\ref{eq:model-1}, the probability of the sequence
$y_{1:T}$ is  defined as the sum of the probabilities of all the
segmentations in $\setS_y$, 
\begin{align}
  p(y_{1:T} |x_{1:T'}) \triangleq \sum_{a_{1:T'} \in S_y}
  \prod_{t=1}^{T'} p(a_t|x_t, \pi(a_{1:t-1})), \label{eq:model-2}
\end{align} 
where $p(a_t|x_t, \pi(a_{1:t-1}))$ is the probability of segment $a_t$
given input element $x_t$ and the concatenation of all previous
segments $\pi(a_{1:t-1})$. In other words, input element $x_t$
emits segment $a_t$. Again this segment probability can be modeled
using an RNN with a softmax probability function with $x_t$ and
$\pi(a_{1:t-1})$ providing the information for the initial state. The
number of possible segments for $y_{1:T}$ is $O(T'T^2)$.  Similar 
to Eq.~\ref{eq:model-1},  a direct computation of Eq.~\ref{eq:model-2} 
is not feasible since $|\setS_y|$ is exponentially large.  
We address the computational details in Section~\ref{sec:alg}. 

\subsection{Carrying over information across segments}
\label{subsec:connecting}
\begin{figure}[t]
\begin{center}
\centerline{\includegraphics[width=0.93\columnwidth]{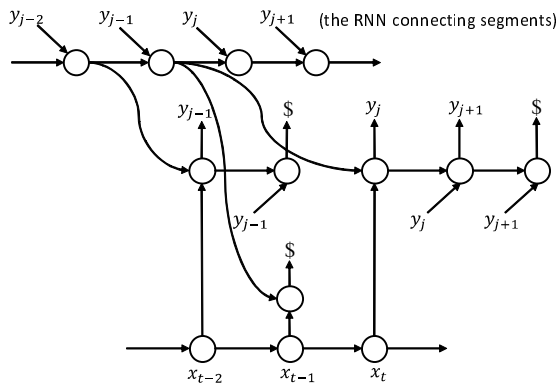}}
\vskip -0.22in
\caption{For Section~\ref{subsec:connecting}. SWAN carries over
  information across segments using a separate RNN. Here the segments
  are $a_{t-2} = \{y_{j-1}, \$\}$, $a_{t-1} = \{\$\}$ and $a_t =
  \{y_j, y_{j+1}, \$\}$ emitted by input elements $x_{t-2}$, $x_{t-1}$ and
$x_t$ respectively.}
\label{fig:wasm-concat}
\end{center}
\vskip -0.25in
\end{figure} 

Note that we do not assume that the segments in a
segmentation are conditionally independent. Take Eq.~\ref{eq:model-2}
as an example, the probability of a segment $a_t$ given $x_t$ is
defined as $p(a_t|x_t, \pi(a_{1:t-1}))$, which also depends on the
concatenation of all previous segments $\pi(a_{1:t-1})$. We take an
approach inspired by the sequence transducer~\cite{graves2012sequence} to
use a separate RNN to model $\pi(a_{1:t-1})$. The hidden state of this
RNN and input $x_t$ are used as the initial state of the RNN
for segment $a_t$. (We simply add them together in our speech recognition experiment.) 
This allows all previous emitted outputs to
affect this segment $a_t$. Figure~\ref{fig:wasm-concat} illustrates
this idea. The significance of this approach is that it still permits the 
exact dynamic programming algorithm as we will describe in Section \ref{sec:alg}.  


\subsection{Related work}
Our approach, especially SWAN, is inspired by 
connectionist temporal classification
(CTC)~\cite{graves2006connectionist} and the sequence
transducer~\cite{graves2012sequence}. CTC defines a 
distribution over the output sequence that is not longer 
than the input sequence. To appropriate map the input to the 
output, CTC marginalizes out all possible alignments using dynamic programming. 
Since CTC does not model the interdependencies
among the output sequence, the sequence transducer introduces
a separate RNN as a prediction network to bring in output-output dependency, 
where the prediction network works like a language model. 


SWAN can be regarded as a generalization of 
CTC to allow segmented outputs. Neither CTC nor the 
sequence transducer takes into account segmental structures of 
output sequences. Instead, our method constructs a probabilistic 
distribution over output sequences by marginalizing all valid segmentations.  
This introduces additional nontrivial computational challenges 
beyond CTC and the sequence transducer. When the input is also a 
sequence, our method then marginalizes the alignments 
between the input and the output segmentations. 
Since outputs are modeled with segmental structures,  
our method can be applied to the scenarios where the input is not a sequence 
or the input length is shorter than the output length,   while CTC cannot.
When we need to carry information across segments, we
borrow the idea of the sequence 
transducer to use a separate RNN. Although it is suspected that using a separate 
RNN could result in a loosely-coupled 
model~\cite{graves2013generating,jaitly2016online} 
that might hinder the performance, we do not find it to be an issue in our approach. 
This is perhaps due to our use of the output segmentation---the hidden states 
of the separate RNN are not directly used for prediction  but as the initial states 
of the RNN for the segments, which strengthens their dependencies
on each other. 

SWAN itself is most 
similar to the recent work on the neural 
transducer~\cite{jaitly2016online}, although we start with a 
different motivation. The motivation of
the neural transducer is to allow incremental predictions as input
streamingly arrives, for example in speech recognition. From the
modeling perspective, it also assumes that the output
is decomposed into several segments and the alignments are unknown in
advance. However, its assumption that hidden states are carried over
across the segments prohibits exact marginalizing all valid
segmentations and alignments. So they resorted to find an approximate ``best''
alignment with a dynamic programming-like algorithm during training or
they might need a separate GMM-HMM model to generate alignments in
advance to achieve better results. Otherwise, without carrying 
information across segments results in sub-optimal performance as shown
in~\citet{jaitly2016online}. In contrast, our method of connecting 
the segments described in Section~\ref{subsec:connecting}
preserves the advantage of exact marginalization over all possible
segmentations and alignments while still allowing the previous emitted
outputs to affect the states of subsequent segments. This allows us to
obtain a comparable good performance without using an additional alignment tool.

Another closely related work is the online segment to segment
neural transduction~\cite{yu2016online}. This work treats the alignments 
between the input and output sequences as latent variables and seeks to 
marginalize them out. From this perspective, SWAN is similar to theirs.  
However, our work explicitly takes into account output segmentations, extending 
the scope of its application to the case when the input is not a sequence. 
Our work is also related  to semi-Markov conditional random 
fields~\cite{Sarawagi04semi-markovconditional}, segmental recurrent neural networks~\cite{kong2015segmental} 
and segmental hidden dynamic model~\cite{DengJaitly2015}, 
where the segmentation is applied to the input sequence instead 
of the output sequence.

\section{Forward, backward and decoding} 
\label{sec:alg}
In this section, we first present the details of forward and backward
computations using dynamic programming. We then describe the beam
search decoding algorithm. With these algorithms, our approach becomes
a standalone loss function that can be used in many
applications. Here we focus on developing the
algorithm for the case when the input is a sequence.  When the input
is not a sequence, the corresponding algorithms can be similarly
derived.

\subsection{Forward and backward propagations}
\paragraph{Forward.}
Consider calculating the result for Eq.~\ref{eq:model-2}. We first
define the forward and backward probabilities,\footnote{The forward and
backward probabilities are terms for dynamic programming and not to
be confused with forward and backward propagations in general
machine learning.}
\begin{align*}
  \alpha_t(j) &= p(y_{1:j} | x_{1:t}), \\
  \beta_t(j) &= p(y_{j+1:T} | x_{t+1:T'}, y_{1:j}),
\end{align*}
where forward $\alpha_t(j)$ represents the probability that input
$x_{1:t}$ emits output $y_{1:j}$ and backward
$\beta_t(j)$ represents the probability that input
$x_{t+1:T'}$ emits output $y_{j+1:T}$. Using
$\alpha_t(j)$ and $\beta_t(j)$, we can verify the following, for any
$t=0,1,...,T'$, 
\begin{align}
  p(y_{1:T} |x_{1:T'}) = \sum_{j=0}^T\alpha_t(j)\beta_t(j),
  \label{eq:dp-result}
\end{align}
where the summation of $j$ from $0$ to $T$ is to enumerate all
possible two-way partitions of output $y_{1:T}$.  A special case is
that $p(y_{1:T} |x_{1:T'}) = \alpha_{T'}(T) = \beta_0(0)$.
Furthermore, we have following dynamic programming recursions
according to the property of the segmentations,
\begin{align}
  \alpha_t(j) & = \sum_{j'=0}^j\alpha_{t-1}(j')p(y_{j'+1:j}|x_t),
  \label{eq:forward}\\
  \beta_t(j) & = \sum_{j'=j}^T\beta_{t+1}(j')p(y_{j+1:j'}|x_{t+1}),
  \label{eq:backward}
\end{align}
where $p(y_{j'+1:j}|x_t)$ is the probability of the segment
$y_{j'+1:j}$ emitted by $x_t$ and $p(y_{j+1:j'}|x_{t+1})$ is similarly
defined. When $j=j'$, notation $y_{j+1:j'}$ indicates an empty
segment with previous output as $y_{1:j}$. For simplicity, we omit
the notation for those previous outputs, since it
does not affect the dynamic programming algorithm. As we discussed
before, $p(y_{j'+1:j}|x_t)$ is modeled using an RNN with a softmax
probability function. Given initial conditions $\alpha_0(0) = 1$ and
$\beta_{T'}(T)=1$, we can efficiently compute the probability of the
entire output $p(y_{1:T} |x_{1:T'})$.

\paragraph{Backward.}
We only show how to compute the gradient w.r.t $x_t$ since others can
be similarly derived. Given the representation of
$p(y_{1:T}|x_{1:T'})$ in Eq.~\ref{eq:dp-result} and the dynamic
programming recursion in Eq.~\ref{eq:forward}, we have
\begin{align}
\frac{\partial \log p(y_{1:T}|x_{1:T'})}{\partial x_t}
=\sum_{j'=0}^T\sum_{j=0}^{j'} w_t(j,j') \frac{\partial \log
p(y_{j+1:j'}|x_t)}{\partial x_t}, \label{eq:grad}
\end{align}
where $w_t(j,j')$ is defined as
\begin{align}
w_t(j,j') \triangleq
{\alpha_{t-1}(j)\beta_t(j')}\frac{p(y_{j+1:j'}|x_t)}{p(y_{1:T}|x_{1:T'})}.
\label{eq:weight}
\end{align}
Thus, the gradient w.r.t. $x_t$ is a weighted linear combination of the contributions from related segments. 

\paragraph{More efficient computation for segment probabilities.}
The forward and backward algorithms above assume that all segment
probabilities, $\log p(y_{j+1:j'}|x_t)$ as well as their gradients
$\frac{\partial \log p(y_{j+1:j'}|x_t)}{\partial x_t}$, for $0\leq j\leq j'\leq T$ 
and $0\leq t\leq T'$, are already computed. There are $O(T'T^2)$ of such
segments. And if we consider each recurrent step as a unit of
computation, we have the computational complexity as $O(T'T^3)$.
Simply enumerating everything, although parallelizable for different
segments, is still expensive.

We employ two additional strategies to allow more efficient
computations. The first is to limit the maximum segment length to be $L$,
which reduces the computational complexity to $O(T'TL^2)$. The second is 
to explore the structure of the segments to further reduce the complexity to $O(T'TL)$. This is an important improvement, 
without which we find the training would be extremely slow.

\begin{figure*}[t]
\begin{center}
\vskip -0.1in
\centerline{\includegraphics[width=0.76\textwidth]{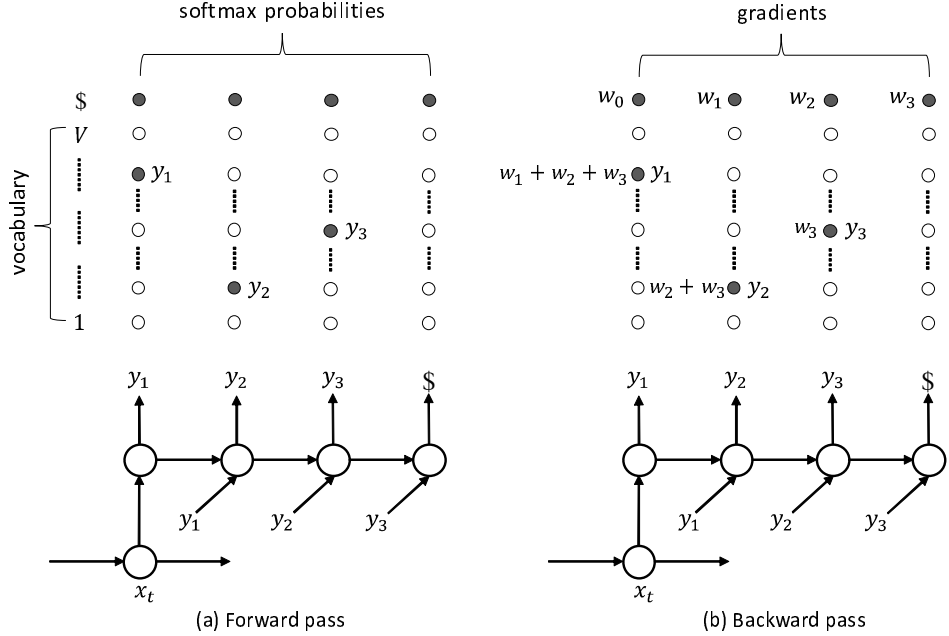}}
\vskip -0.2in
\caption{Illustration for an efficient computation for segments
  $y_{j+1:j'}$, $j'=j, j+1, ..., j+L$  with one pass on the longest
  segment $y_{j+1:j+L}$, where $V$ is the vocabulary size and $\$$ is
  the symbol for the end of a segment.  In this example, we use $j=0$ and
  $L=3$. Thus we have four possible segments $\{\$\}$, $\{y_1, \$\}$,
  $\{y_1, y_2, \$\}$ and $\{y_1, y_2, y_3, \$\}$ given input $x_t$.
  (a) Forward pass. Shaded small circles indicate the softmax
  probabilities needed to compute the probabilities of all four
  segments. (b) Backward pass.  The weights are $w_{j'} \triangleq
  w_t(0,j')$ defined in~Eq.\ref{eq:weight} for $j'=0,1,2,3$ for four
  segments mentioned above. Shaded small circles are annotated with
  the gradient values while unshaded ones have zero gradients. For
example, $y_1$ has a gradient of $w_1+w_2+w_3$ since $y_1$ appears in
three segment $\{y_1, \$\}$, $\{y_1, y_2, \$\}$ and $\{y_1, y_2, y_3,
\$\}$.}
\label{fig:grad}
\end{center}
\vskip -0.25in
\end{figure*}

The key observation for the second strategy is that the computation
for the longest segment can be used to cover those for the shorter
ones.  First consider forward propagation with $j$ and $t$ fixed. 
Suppose we want to compute
$\log p(y_{j+1:j'}|x_t)$ for any $j'=j,..., j+L$, which contains
$L+1$ segments, with the length ranging from $0$ to $L$. In order to
compute for the longest segment $\log p(y_{j+1:j+L}|x_t)$, we need the
probabilities for $p(y=y_{j+1}|x_t, h_0)$, $p(y=y_{j+2}|y_{j+1}, x_t,
h_1)$, ..., $p(y=y_{j+L}|y_{j+L-1}, x_t, h_{L-1})$ and
$p(y=\$|y_{j+L}, x_t, h_L)$, where $h_l$, $l = 0,1,...,L$, are the
recurrent states. Note that this process also gives us the probability
distributions needed for the shorter segments when $j'=j,..., 
j+L-1$. For backward propagation, we observe that, from
Eq.~\ref{eq:grad}, each segment has its own weight on the contribution
to the gradient, which is $w_t(j, j')$ for $p(y_{j+1:j'}|x_t)$, $j'=j,..., j+L$.  
Thus all we need is to assign proper weights
to the corresponding gradient entries for the longest segment 
$y_{j+1:j+L}$ in order to integrate the contributions from 
the shorter ones. Figure~\ref{fig:grad} illustrates the forward 
and backward procedure.


\subsection{Beam search decoding}
Although it is possible compute the output sequence probability using 
dynamic programming during training, it is impossible to do a similar 
thing during decoding since the output is unknown. 
We thus resort to beam search. The beam search
for SWAN is more complex than the simple left-to-right beam search
algorithm used in standard sequence-to-sequence
models~\cite{Sutskever:2014}. In fact, for each input element $x_t$,
we are doing a simple left-to-right beam search decoder. In addition,
different segmentations might imply the same output sequence and we
need to incorporate this information into beam search as well. To
achieve this, each time after we process an input element $x_t$, we
merge the partial candidates with different segments into one
candidate if they indicate the same partial sequence. This is
reasonable because the emission of the next input element $x_{t+1}$ only
depends on the concatenation of all previous segments as discussed in
Section~\ref{subsec:connecting}. Algorithm~\ref{alg:wasn-decoding}
shows the details of the beam search decoding algorithm.
\begin{algorithm}[tb]
\begin{small}
  \caption{SWAN beam search decoding}
   \label{alg:wasn-decoding}
\begin{algorithmic}
  \STATE {\bfseries Input:} input $x_{1:T'}$, beam size $B$, maximum
segment length $L$, $\mathcal{Y} = \{\varnothing\}$ and $\mathcal{P} = \{\varnothing:1\}$.
   \FOR{$t=1$ {\bfseries to} $T'$}
   \STATE {// A left-to-right beam search given $x_t$.}
   \STATE {Set local beam size $b = B$, $\mathcal{Y}_t = \{\}$ and $\mathcal{P}_t = \{\}$.}
   \FOR{$j=0$ {\bfseries to} $L$}
   \FOR {${\bf y} \in \mathcal{Y}$} 
   \STATE {Compute the distribution of the next output for current
   segment, $p(y_j | {\bf y}, x_t)$.}
   \ENDFOR
   \IF{$j = L$}
   \STATE {// Reaching the maximum segment length.}
   \FOR{${\bf y} \in \mathcal{Y}$}
   \STATE {$\mathcal{P}({\bf y}) \leftarrow \mathcal{P}({\bf y})p(y_j =\$| {\bf y}, x_t)$}
   \ENDFOR
   \STATE {Choose $b$ candidates with highest probabilities
   $\mathcal{P}({\bf y})$ from $\mathcal{Y}$ and move them into
 $\mathcal{Y}_t$ and $\mathcal{P}_t$.}
   \ELSE
   \STATE {Choose a set $\mathcal{Y}_{\rm tmp}$ containing $b$
   candidates with highest probabilities $\mathcal{P}({\bf y})p(y_j |
 {\bf y}, x_t)$ out of all pairs $\{{\bf y}, y_j\}$, where ${\bf y}
 \in \mathcal{Y}$ and $y_i \in \{1,...,V, \$\}$}.
 \FOR{$\{{\bf y}, y_j\} \in \mathcal{Y}_{\rm tmp}$}
 \STATE{$\mathcal{P}({\bf y})\leftarrow\mathcal{P}({\bf y})p(y_j |
 {\bf y}, x_t)$.}
 \IF {$y_j = \$$}
 \STATE {Move ${\bf y}$ from $\mathcal{Y}$ and $\mathcal{P}$ into  $\mathcal{Y}_t$ and $\mathcal{P}_t$.}
 \STATE {$b \leftarrow b - 1$.}
 \ELSE
 \STATE{${\bf y} \leftarrow \{{\bf y}, y_j\}$ }
 \ENDIF
  \ENDFOR
   \ENDIF
   \IF {$b = 0$}
   \STATE{break}
   \ENDIF
   \ENDFOR
   \STATE {Update $\mathcal{Y} \leftarrow \mathcal{Y}_t$ and
   $\mathcal{P} \leftarrow \mathcal{P}_t$.}
   \STATE {// Merge duplicate candidates in ${\mathcal{Y}}$.} 
     \WHILE {There exists ${\bf y}_i = {\bf y}_{i'}$ for any ${\bf y}_i, {\bf
       y}_{i'}  \in \mathcal{Y}$}
       \STATE {$\mathcal{P}({\bf y}_i) \leftarrow \mathcal{P}({\bf y}_i) + \mathcal{P}({\bf y}_{i'})$}
       \STATE {Remove ${\bf y}_{i'}$ from ${\mathcal{Y}}$ and ${\mathcal{P}}$.}
     \ENDWHILE
   \ENDFOR
   \STATE {\bfseries Return:} output ${\bf y}$ with the highest
   probability from $\mathcal{Y}$.
\end{algorithmic}
\end{small}
\end{algorithm}

\section{Experiments}
\label{sec:exp}
In this section, we apply our method to two applications, one
unsupervised and the other supervised. These include 1) content-based
text segmentation, where the input to our distribution is a vector 
(constructed using a variational autoencoder for text) and
2) speech recognition, where the input to our distribution
is a sequence (of acoustic features).

\subsection{Content-based text segmentation}
This text segmentation task corresponds to an application of a
simplified version of the non-sequence-input model in
Section~\ref{subsec:non-seq}, where we drop the term $\pi(a_{1:t-1})$
in Eq.\ref{eq:model-1}.

\paragraph{Model description.}
In this task, we would like to automatically discover segmentations
for textual content. To this end, we build a simple model inspired by
latent Dirichlet allocation (LDA)~\cite{Blei:2003b} and neural
variational inference for texts~\cite{miao2016neural}.

LDA assumes that the words are exchangeable within a document---``bag
of words'' (BoW). We generalize this assumption to the segments within
each segmentation---``bag of segments''.  In other words, if we had a
pre-segmented document, all segments would be exchangeable.  However,
since we do not have a pre-segmented document, we assume that for any
valid segmentation. In addition, we choose to drop the term $\pi(a_{1:t-1})$
in Eq.\ref{eq:model-1} in our sequence distribution so that 
we do not carry over information across segments. Otherwise, 
the segments are not exchangeable. This is designed to be comparable 
with the exchangeability assumption in LDA,
although we can definitely use the carry-over technique in other
occasions.

Similar to LDA, for a document with words $y_{1:T}$, we assume that a
topic-proportion like vector, $\theta$, controls the distribution of
the words. In more details, we define $\theta (\zeta) \propto \exp(\zeta)$,
where $\zeta \sim \mathcal{N}(0, I)$.  Then the log likelihood of
words $y_{1:T}$ is defined as
\begin{align*}
  \log p(y_{1:T}) &= \log \mathbb{E}_{p(\zeta)} [
p(y_{1:T}|W\theta(\zeta))]
\\ & \textstyle \geq  \mathbb{E}_{q(\zeta)} [\log p(y_{1:T}|W\theta(\zeta))] +
\mathbb{E}_{q(\zeta)}\left[\log \frac{p(\zeta)} {q(\zeta)}\right],
\end{align*} 
where the last inequality follows the variational inference
principle~\cite{Jordan:1999a} with variational distribution
$q(\zeta)$. Here $p(y_{1:T}|W\theta(\zeta))$ is modeled as
Eq.\ref{eq:model-1} with $W\theta(\zeta)$ as the input 
vector ``$x$'' and $W$ being another weight matrix. Note again
that $\pi(a_{1:t-1})$ is not used in $p(y_{1:T}|W\theta(\zeta))$.

For variational distribution $q(\zeta)$, we use variational
autoencoder to model it as an inference
network~\cite{Kingma:2013,rezende2014stochastic}. We use the form
similar to~\citet{miao2016neural}, where the inference network is a
feed-forward neural network and its input is the BoW of the document ---
$q(\zeta) \triangleq q(\zeta|{\rm BoW}(y_{1:T}))$.

\paragraph{Predictive likelihood comparison with LDA.} We use two
datasets including AP (Associated Press, $2,246$ documents)
from~\citet{Blei:2003b} and
CiteULike\footnote{\url{http://www.citeulike.org}} scientific article
abstracts ($16,980$ documents) from~\citet{Wang:2011}. Stop words are
removed and a vocabulary size of $10,000$ is chosen by tf-idf for both
datasets. Punctuations and stop words are considered to be known
segment boundaries for this experiment. 
For LDA, we use the variational EM implementation taken from authors'
website.\footnote{\url{http://www.cs.columbia.edu/~blei/lda-c/}} 

We vary the number of topics to be $100$, $150$, $200$, $250$ and $300$. And
we use a development set for early stopping with up to 100 epochs. For
our model, the inference network is a 2-layer feed-forward neural
network with ReLU nonlinearity. A two-layer GRU is used to model the
segments in the distribution $p(y_{1:T}|W\theta(\zeta))$. And we vary
the hidden unit size (as well as the word embedding size) to be $100$,
$150$ and $200$, and the maximum segment length $L$ to be $1$, $2$ and
$3$.  We use Adam algorithm~\cite{kingma2014adam} for optimization
with batch size $32$ and learning rate $0.001$.

We use the evaluation setup from~\citet{Hoffman:2013} for comparing two
different models in terms of predictive log likelihood on a heldout set. 
We randomly choose $90\%$ of documents for training and
the rest is left for testing. For each document ${\bf y}$ in testing,
we use first $75\%$ of the words, ${\bf y}_{obs}$, for estimating
$\theta(\zeta)$ and the rest, ${\bf y}_{eval}$, for evaluating the
likelihood. We use the mean of $\theta$ from variational distribution
for LDA or the output of inference network for our model. For our
model, $p({\bf y}_{eval}| {\bf y}_{obs}) \approx p({\bf y}_{eval}| W\theta(\bar{\zeta}_{obs}))$, 
where $\bar{\zeta}_{obs}$ is chosen as the mean of $q(\zeta| {\bf
y}_{obs})$.  Table~\ref{tab:likelihood} shows the empirical results.
When the maximum segment length $L=1$, our model is better on AP
but worse on CiteULike than LDA. When $L$ increases from 1 to 2 and 3, 
our model gives monotonically higher predictive likelihood on both datasets, 
demonstrating that bringing in segmentation information leads to a better model.
\begin{table}[t]
\caption{Predictive log likelihood comparison. Higher values indicate better results. 
$L$ is the maximum segment length. The top table shows LDA results and the bottom one shows ours.}
\label{tab:likelihood}
\vskip 0.1in
\begin{center}
\begin{small}
\begin{sc}
\begin{tabular}{ccc}
\hline
\abovespace\belowspace
\#LDA Topics & AP & CiteULike \\
\hline
\abovespace
100   & -9.25 &  -7.86 \\
150   & -9.23 & -7.85  \\
200   &\textbf{-9.22} & -7.83  \\
250   & -9.23 & \textbf{-7.82}  \\
\belowspace
  300    & \textbf{-9.22} & \textbf{-7.82}  \\
\hline
\end{tabular}
\vskip 0.2in
\begin{tabular}{cccc}
\hline
\abovespace\belowspace
\#Hidden & $L$ &  AP & CiteULike \\
\hline
\abovespace
  100   & 1 & -8.42  & -8.12 \\
  100   & 2 &  -8.31 & -7.68 \\
\belowspace
  100   & 3 & -8.29  & -7.61  \\
\hline
\abovespace
  150   & 1 & -8.38 & -8.12 \\
  150   & 2 & -8.30 & -7.67 \\
\belowspace
  150   & 3 & \textbf{-8.28} & \textbf{-7.60} \\
\hline
\abovespace
  200    & 1 & -8.41  & -8.13 \\
  200    & 2 & -8.32 & -7.67 \\
\belowspace
  200    & 3 & -8.30 & -7.61 \\
\hline
\end{tabular}
\end{sc}
\end{small}
\end{center}
\vskip -0.25in
\end{table}

\begin{figure}[t]
\begin{center}
\centerline{\includegraphics[width=1.05\columnwidth]{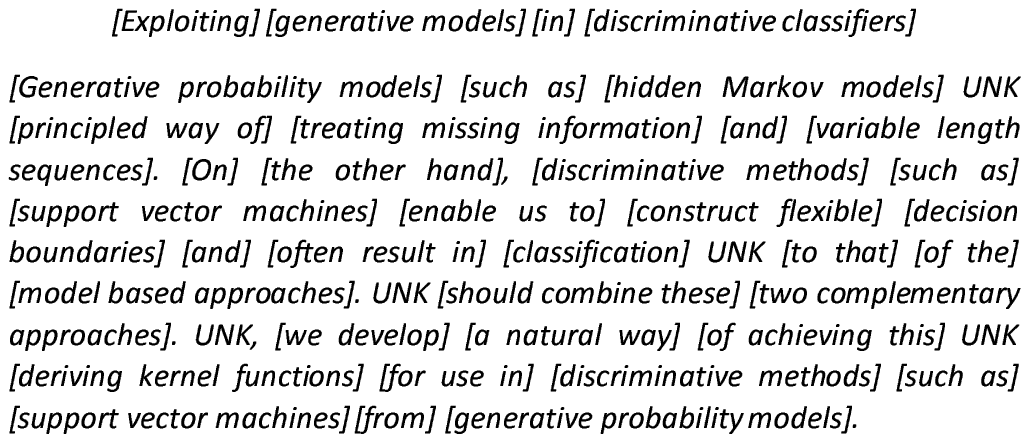}}
\vskip -0.15in
\caption{Example text with automatic segmentation, 
obtained using the path with highest probability. Words in the same
brackets ``[ ]'' belong to the same segment. ``UNK'' indicates a
word not in the vocabulary. The maximum segment length $L=3$.}
\label{fig:example-seg} \end{center}
\vskip -0.2in
\end{figure}

\paragraph{Example of text segmentations.} In order to improve the
readability of the example segmentation, we choose to keep the stop
words in the vocabulary, different from the setting in the
quantitative comparison with LDA. Thus, stop words are not treated as
boundaries for the segments. Figure~\ref{fig:example-seg} shows an
example text. The segmentation is obtained by finding the path with
the highest probability in dynamic programming.\footnote{This is done
 by replacing the ``sum'' operation with ``max'' operation in
 Eq.~\ref{eq:forward}.} As we can see, many reasonable segments are
 found using this automatic procedure. 

\subsection{Speech recognition}
\label{subsec:speech}
We also apply our model to speech recognition, and present results 
on both phoneme-level and character-level experiments. This
corresponds to an application of SWAN described in
Section~\ref{subsec:seq}.

\paragraph{Dataset.}
We evaluate SWAN on the TIMIT corpus following the setup in~\citet{Deng2006}. 
The audio data is encoded using a Fourier-transform-based filter-bank with 40
coefficients (plus energy) distributed on a mel-scale, together with
their first and second temporal derivatives. Each input vector is
therefore size 123. The data is normalized so that every element of
the input vectors has zero mean and unit variance over the training
set.  All 61 phoneme labels are used during training and decoding,
then mapped to 39 classes for scoring in the standard way~\cite{lee1989speaker}.

\paragraph{Phoneme-level results.}
Our SWAN model consists of a $5$-layer bidirectional GRU with 300 hidden
units as the encoder and two $2$-layer unidirectional GRU(s) with 600
hidden units, one for the segments and the other for connecting the
segments in SWAN.  We set the maximum segment length $L=3$. To reduce
the temporal input size for SWAN, we add a temporal convolutional
layer with stride $2$ and width $2$ at the end of the encoder. For
optimization, we largely followed the strategy
in~\citet{zhang2017towards}. We use Adam \cite{kingma2014adam} with
learning rate $4e-4$. We then use stochastic gradient descent with
learning rate $3e-5$ for fine-tuning.  Batch size $20$ is used during
training. We use dropout with probability of $0.3$ across the layers
except for the input and output layers.  Beam size $40$ is used for
decoding. Table~\ref{tab:timit_per} shows the results compared with
some previous approaches. SWAN achieves competitive results 
without using a separate alignment tool.

We also examine the properties of SWAN's outputs.
We first estimate the {\it average segment length}\footnote{The average
segment length is defined as the length of the output (excluding 
end of segment symbol $\$$) divided by the number of segments 
(not counting the ones only containing $\$$).} $\ell$ for the output. 
We find that $\ell$ is usually smaller than $1.1$ 
from the settings with good performances. 
Even when we increase the maximum segment length 
$L$ to $6$, we still do not see a significantly 
increase of the average segment length. We suspect that
the phoneme labels are relatively independent summarizations of the
acoustic features and it is not easy to find good phoneme-level
segments. The most common segment patterns we observe are `sil ?', 
where `sil' is the silence phoneme label and `?' denotes
some other phoneme label~\cite{lee1989speaker}. 
On running time, SWAN is about 5 times slower than CTC. 
(Note that CTC was written in CUDA C, while SWAN is written in torch.)

\begin{table*}[th!]
  \caption{Examples of character-level outputs with their segmentations, 
  where ``$\cdot$'' represents the segment boundary, ``$\square$'' represents the
    space symbol in SWAN's outputs, the ``best path'' represents the
    most probable segmentation given the ground truth, and the ``max
  decoding'' represents the beam search decoding result with beam size
1.} \label{tab:character-level}
	\vskip 0.15in
	{\small 
		\begin{center}
			\begin{tabular}{rl}
				\hline
\abovespace
				ground truth& one thing he thought nobody knows about it yet\\
				best path&o$\cdot$ne$\square$$\cdot$th$\cdot$i$\cdot$ng$\square$$\cdot$he$\square$$\cdot$th$\cdot$ou$\cdot$ght$\cdot$$\square$$\cdot$n$\cdot$o$\cdot$bo$\cdot$d$\cdot$y$\square$$\cdot$kn$\cdot$o$\cdot$w$\cdot$s$\square$$\cdot$a$\cdot$b$\cdot$ou$\cdot$t$\square$$\cdot$i$\cdot$t$\square$$\cdot$y$\cdot$e$\cdot$t\\
\belowspace
				max decoding&o$\cdot$ne$\square$$\cdot$th$\cdot$a$\cdot$n$\cdot$$\square$$\cdot$he$\square$$\cdot$th$\cdot$ou$\cdot$gh$\cdot$o$\cdot$t$\square$$\cdot$n$\cdot$o$\cdot$bo$\cdot$d$\cdot$y$\square$$\cdot$n$\cdot$o$\cdot$se$\square$$\cdot$a$\cdot$b$\cdot$ou$\cdot$t$\square$$\cdot$a$\cdot$t$\square$$\cdot$y$\cdot$e$\cdot$t\\
  \hline
  \abovespace
				ground truth& jeff thought you argued in favor of a centrifuge purchase\\
				best path&j$\cdot$e$\cdot$ff$\cdot$$\square$$\cdot$th$\cdot$ou$\cdot$ght$\square$$\cdot$you$\cdot$$\square$$\cdot$a$\cdot$r$\cdot$g$\cdot$u$\cdot$ed$\square$$\cdot$in$\square$$\cdot$f$\cdot$a$\cdot$vor$\cdot$$\square$$\cdot$of$\square$$\cdot$a$\cdot$$\square$$\cdot$c$\cdot$en$\cdot$tr$\cdot$i$\cdot$f$\cdot$u$\cdot$ge$\square$$\cdot$p$\cdot$ur$\cdot$ch$\cdot$a$\cdot$s$\cdot$e\\
\belowspace
				max decoding&j$\cdot$a$\cdot$ff$\cdot$$\square$$\cdot$th$\cdot$or$\cdot$o$\cdot$d$\cdot$y$\square$$\cdot$a$\cdot$re$\square$$\cdot$g$\cdot$i$\cdot$vi$\cdot$ng$\square$$\cdot$f$\cdot$a$\cdot$ver$\cdot$$\square$$\cdot$of$\square$$\cdot$er$\cdot$s$\cdot$e$\cdot$nt$\cdot$$\square$$\cdot$f$\cdot$u$\cdot$ge$\square$$\cdot$p$\cdot$er$\cdot$ch$\cdot$e$\cdot$s\\
  \hline
  \abovespace
				ground truth& he trembled lest his piece should fail\\
				best path&he$\cdot$$\square$$\cdot$tr$\cdot$e$\cdot$m$\cdot$b$\cdot$le$\cdot$d$\square$$\cdot$l$\cdot$e$\cdot$s$\cdot$t$\cdot$$\square$$\cdot$hi$\cdot$s$\square$$\cdot$p$\cdot$i$\cdot$e$\cdot$ce$\square$$\cdot$sh$\cdot$oul$\cdot$d$\square$$\cdot$f$\cdot$a$\cdot$i$\cdot$l\\
\belowspace
				max decoding&he$\cdot$$\square$$\cdot$tr$\cdot$e$\cdot$m$\cdot$b$\cdot$le$\square$$\cdot$n$\cdot$e$\cdot$s$\cdot$t$\cdot$$\square$$\cdot$hi$\cdot$s$\square$$\cdot$p$\cdot$ea$\cdot$s$\cdot$u$\cdot$de$\square$$\cdot$f$\cdot$a$\cdot$i$\cdot$l\\
  \hline
			\end{tabular}
		\end{center}
	}
	\vskip -0.3in
\end{table*}

 \begin{table}[t]
 	\vskip 0.1in
  \caption{TIMIT phoneme recognition results. ``PER'' is the phoneme
  error rate on the core test set.}
 	\label{tab:timit_per}
 	\vskip 0.1 in
 	{\small 
 	\begin{center}
 	
 				\begin{tabular}{cc}
 					\hline
\abovespace\belowspace
 					Model & PER (\%) \\\hline
\abovespace\
 					BiLSTM-5L-250H \cite{graves2013speech}  &  18.4 \\
 					TRANS-3L-250H \cite{graves2013speech}  & 18.3 \\
                    Attention RNN \cite{chorowski2015attention} & 17.6 \\
 					Neural Transducer \cite{jaitly2016online}   & 18.2 \\
 					CNN-10L-maxout \cite{zhang2017towards}   & 18.2 \\
 					\belowspace
 					SWAN (this paper) & 18.1 \\\hline
 				\end{tabular}
	\end{center}
	}
 	\vskip -0.25in
 \end{table}
\paragraph{Character-level results.}
In additional to phoneme-level recognition experiments, we also
evaluate our model on the task to directly output the characters
like~\citet{amodei2016deep}. We use the original word level
transcription from the TIMIT corpus, convert them into lower cases,
and separate them to character level sequences (the vocabulary
includes from `a' to `z', apostrophe and the space symbol.)  We find
that using temporal convolutional layer with stride $7$ and width $7$
at the end of the decoder and setting $L=8$ yields good results. 
In general, we found that starting with a larger $L$ is useful.
We believe that a larger $L$ allows more explorations of different segmentations 
and thus helps optimization since we consider the marginalization of 
all possible segmentations. We obtain a character error rate (CER) 
of ${\bf 30.5\%}$ for SWAN compared
to $31.8\%$ for CTC.\footnote{As far as we know, there is no public
CER result of CTC for TIMIT, so we empirically find 
the best one as our baseline. We use Baidu's CTC implementation: 
\url{https://github.com/baidu-research/warp-ctc}.}

\begin{figure}[t]
\begin{center}
	\includegraphics[width=1.0\linewidth]{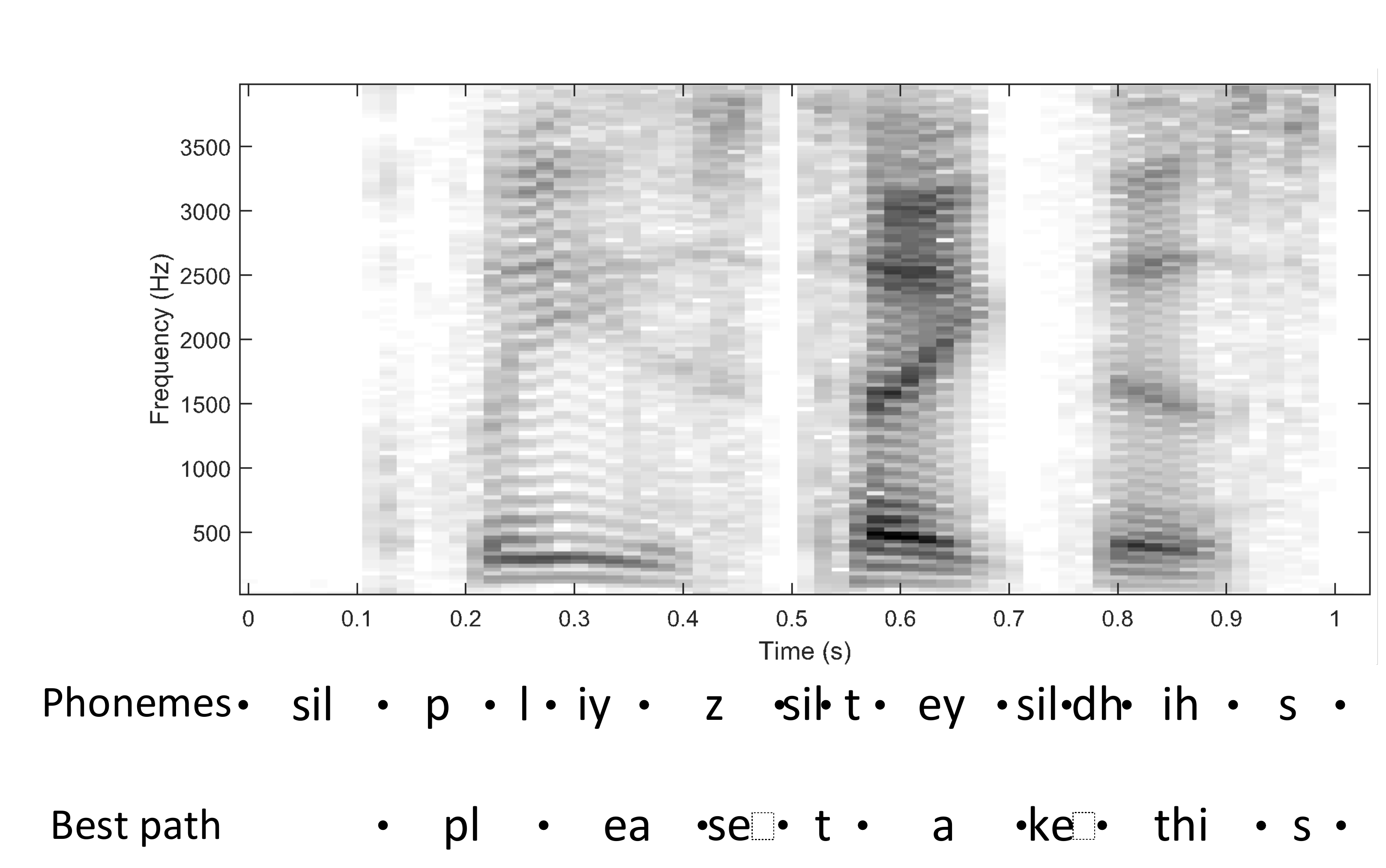}
  \caption{Spectrogram of a test example of the output sequence, ``please
    take this''. Here ``$\cdot$'' represents the boundary and
    $\square$ represents the space symbol in SWAN's result. The
    ``phonemes'' sequence is the ground truth phoneme labels. 
    (The full list of phoneme labels and their explanations can be found 
    in~\citet{lee1989speaker}.) The ``best path'' sequence is from SWAN. 
    Note that the time boundary is not precise due to the convolutional layer.}
\label{fig:spectrogram}
\end{center}
 	\vskip -0.25in
\end{figure}

We examine the properties of SWAN for this
character-level recognition task.  Different from the
observation from the phoneme-level task, we find the average segment
length $\ell$ is around $1.5$ from the settings with good performances,
longer than that of the phoneme-level setting. This is expected since the 
variability of acoustic features for a character 
is much higher than that for a phone and a longer segment of characters 
helps reduce that variability.
Table~\ref{tab:character-level} shows some example decoding outputs.
As we can see, although not perfect, these segments often correspond to 
important phonotactics rules in the English language 
and we expect these to get better 
when we have more labeled speech data. In Figure~\ref{fig:spectrogram}, 
we show an example of mapping the character-level alignment back to the
speech signals, together with the ground truth phonemes. 
We can observe that the character level sequence roughly
corresponds to the phoneme sequence in terms of phonotactics 
rules.

Finally, from the examples in Table~\ref{tab:character-level}, we find
that the space symbol is often assigned to a segment together with its
preceding character(s) or as an independent segment. We suspect this is
because the space symbol itself is more like a separator of segments
than a label with actual acoustic meanings. So in future work, we plan
to treat the space symbol between words as a known segmentation
boundary that all valid segmentations should comply with, 
which will lead to a smaller set of possible segments.
We believe this will not only make it easier to 
find appropriate segments, but also significantly reduce the computational complexity.


 
%

\section{Conclusion and Future work} 
In this paper, we present a new probability distribution for
sequence modeling and demonstrate its usefulness on two different
tasks. Due to the generality, it can be used as a loss function in many
sequence modeling tasks. We plan to investigate following directions in future
work. The first is to validate our approach on large-scale speech datasets. The
second is machine translation, where segmentations can be regarded as
``phrases.'' We believe this approach has the potential to bring together the
merits of traditional phrase-based translation~\cite{koehn2003statistical} and
recent neural machine translation~\cite{Sutskever:2014,Bahdanau:2014}. 
For example, we can restrict the number of valid segmentations with a known phrase set.
Finally, applications in other domains including DNA sequence
segmentation~\cite{braun1998statistical} might benefit from our
approach as well.


\bibliography{bib}
\bibliographystyle{icml2017}

\end{document}